# Towards Neural Co-Processors for the Brain: Combining Decoding and Encoding in Brain-Computer Interfaces

Rajesh P. N. Rao
Center for Neurotechnology
Paul G. Allen School of Computer Science and Engineering
University of Washington, Seattle

## Abstract

The field of brain-computer interfaces is poised to advance from the traditional goal of controlling prosthetic devices using brain signals to combining neural decoding and encoding within a single neuroprosthetic device. Such a device acts as a "co-processor" for the brain, with applications ranging from inducing Hebbian plasticity for rehabilitation after brain injury to reanimating paralyzed limbs and enhancing memory. We review recent progress in simultaneous decoding and encoding for closed-loop control and plasticity induction. To address the challenge of multi-channel decoding and encoding, we introduce a unifying framework for developing brain co-processors based on artificial neural networks and deep learning. These "neural co-processors" can be used to jointly optimize cost functions with the nervous system to achieve desired behaviors ranging from targeted neuro-rehabilitation to augmentation of brain function.

## Introduction

A brain-computer interface (BCI) [1,2,3,4] is a device that can (a) allow signals from the brain to be used to control devices such as prosthetics, cursors or robots, and (b) allow external signals to be delivered to the brain through neural stimulation. The field of BCIs has made enormous strides in the past two decades. The genesis of the field can be traced to early efforts in the 1960s by neuroscientists such as Eb Fetz [5] who studied operant conditioning in monkeys by training them to control the movement of a needle in an analog meter by modulating the firing rate of a neuron in their motor cortex. Others such as Delgado and Vidal explored techniques for neural decoding and stimulation in early versions of neural interfaces [6,7]. After a promising start, there was a surprising lull in the field until the 1990s when, spurred by the advent of multi-electrode recordings as well as fast and cheap computers, the field saw a resurgence under the banner of brain-computer interfaces (BCIs; also known as brain-machine interfaces and neural interfaces) [1,2].

A major factor in the rise of BCIs has been the application of increasingly sophisticated machine learning techniques for decoding neural activity for controlling prosthetic arms [8,9,10], cursors [11,12,13,14,15,16], spellers [17,18] and robots [19,20,21,22]. Simultaneously, researchers have explored how information can be biomimetically or artificially encoded and delivered via stimulation to neuronal networks in the brain and other regions of the nervous system for auditory [23], visual [24], proprioceptive [25], and tactile [26,27,28,29,30] perception.

Building on these advances in neural decoding and encoding, researchers have begun to explore bi-directional BCIs (BBCIs) which integrate decoding and encoding in a single system. In this article, we review how BBCIs can be used for closed-loop control of prosthetic devices, reanimation of paralyzed limbs, restoration of sensorimotor and cognitive function, neuro-rehabilitation, enhancement of memory, and brain augmentation. Motivated by this recent progress, we propose a new unifying framework for combining decoding and encoding based on "neural co-processors" which rely on artificial neural networks and deep learning. We show that these "neural co-processors" can be used to jointly optimize cost functions with the nervous system to achieve goals such as targeted rehabilitation and augmentation of brain function, besides providing a new tool for testing computational models and understanding brain function [31].

**Simultaneous Decoding and Encoding in BBCIs**

*Closed-Loop Prosthetic Control*

Consider the problem of controlling a prosthetic hand using brain signals. This involves (1) using recorded neural responses to control the hand, (2) stimulating somatosensory neurons to provide tactile and proprioceptive feedback, and (3) ensuring that stimulation artifacts do not corrupt the recorded signals being used to control the hand. Several artifact reduction methods have been proposed for (3) – we refer the reader to [32,33,34]. We focus here on combining (1) decoding with (2) encoding.

Most state-of-the-art decoding algorithms for intracortical BCIs are based on a linear decoder such as the Kalman filter. Typically, the state vector $x$ for the Kalman filter is chosen to be a vector of kinematic quantities to be estimated, such as hand position, velocity, and acceleration. The likelihood (or measurement) model for the Kalman filter specifies how the kinematic vector $x_t$ at time $t$ relates linearly (via a matrix $B$) to the measured neural activity vector $y_t$:

$$y_t = Bx_t + m_t$$

while a dynamics model specifies how $x_t$ linearly changes (via matrix $A$) over time:

$$x_t = Ax_{t-1} + n_t$$

$n_t$ and $m_t$ are zero-mean Gaussian noise processes. The Kalman filter computes the optimal estimates for kinematics $x_t$ (both mean and covariance) given current and all past neural measurements.

One of the first studies to combine decoding and encoding was by O'Doherty, Nicolelis, and colleagues [35] who showed that stimulation of somatosensory cortex could be used to instruct a rhesus monkey which of two targets to move a cursor to; the cursor was subsequently controlled using a BCI based on linear decoding to predict the X- and Y-coordinate of the cursor. A later study by the same group [36] demonstrated true closed-loop control. Monkeys used a BCI based on primary motor cortex (M1) recordings and Kalman-filter-based decoding to actively explore virtual objects on a screen with artificial tactile properties. The monkeys were rewarded if they found the object with particular artificial tactile properties. During brain-controlled exploration of an object, the associated tactile information was delivered to somatosensory cortex (S1) via intracortical stimulation. Tactile information was encoded as a high-frequency biphasic pulse train (200 Hz for rewarded object, 400 Hz for others) presented in packets at a lower frequency (10 Hz for rewarded, 5 Hz for unrewarded objects). Because stimulation artifacts masked neural activity for 5–10 ms after each pulse, an interleaved scheme of alternating 50 ms recording and 50 ms stimulation was used. The monkeys were able to select the desired target object within a second or less based only on its tactile properties as conveyed through stimulation.

Klaes, Andersen and colleagues [37] have also demonstrated that a monkey can utilize intracortical stimulation in S1 to perform a match-to-sample task where the goal is to move a virtual arm and find a target object that delivers stimulation similar to a control object. In their experiment, the monkey controlled a virtual arm using a Kalman-filter-based decoding scheme where the Kalman filter's state was defined as the virtual hand's position, velocity and acceleration in three dimensions. The encoding algorithm involved stimulating S1 via three closely located electrodes using a 300 Hz biphasic pulse train for up to 1 second while the virtual hand held the object. After training, the monkey was able to move the virtual hand to the correct target with success rates between 70% and more than 90% over the course of 8 days (chance level was 50%).

Finally, Flesher and colleagues [38] have recently shown that a paralyzed patient can use a bidirectional BCI for closed-loop control of a prosthetic hand in a continuous force matching task. Control signals were decoded from multi-electrode recordings in M1 using a linear decoder that mapped M1 firing rates to movement velocities of the robotic arm. Initial training data for the linear decoder was obtained by asking the subject to observe the robotic hand performing hand shaping tasks such as "pinch" (thumb/index/middle flexion-extension), "scoop" (ring/pinky flexion/extension) or grasp (all finger flexion) and recording M1 firing rates, followed by a second training phase involving computer-assisted control to fine tune the decoder weights. The subject then performed a 2D force matching task with the robotic hand using the trained decoder to pinch, scoop or grasp a foam object either gently or firmly while using stimulation of S1 to get feedback on the force applied. The encoding algorithm linearly mapped torque sensor data from the robotic hand's finger motors to pulse train amplitude of those stimulating electrodes that previously elicited percepts on the corresponding fingers of the subject. The researchers showed that the subject was able to continuously control the flexion/extension of the pinch and scoop dimensions while evaluating the applied torque based on force feedback from S1 stimulation. The success rate for pinch, scoop or grasp with gentle or firm forces) was significantly higher with stimulation feedback compared to feedback from vision alone.

*Reanimating Paralyzed Limbs*
Rather than controlling a prosthetic limb, BBCIs can also be used to control electrical stimulation of muscles to restore movement in a paralyzed limb. Moritz, Perlmutter, and Fetz [39] demonstrated this approach in two monkeys by translating the activity of single motor cortical neurons into electrical stimulation of wrist muscles to move a cursor on a computer screen. The decoding scheme involved operant conditioning to volitionally control activity of a motor cortical neuron to initially move a cursor into a target. After training, the activity from the motor cortical neuron was converted into electrical stimuli which was delivered to the monkey's temporarily paralyzed wrist muscles (this type of stimulation is called functional electrical stimulation, or FES). Flexor FES current was set to be proportional to the rate above a threshold (0.8 x [firing rate – 24] with a maximum of 10 mA), and extensor FES was inversely proportional to the rate below a second threshold (0.6 x [12 – firing rate] with a maximum of 10 mA). Both monkeys were able to modulate the activity of cortical neurons to control their paralyzed wrist muscles and move a manipulandum to acquire five targets. Ethier *et al.* [40] extended these results to grasping and moving objects using a linear decoder with a static nonlinearity applied to about 100 neural signals from M1.

Extending the approach to humans, Bouton *et al.* [41] showed that a quadraplegic man with a 96-electrode array implanted in the hand area of the motor cortex could use cortical signals to electrically stimulate muscles in his paralyzed forearm and produce six different wrist and hand motions. For decoding, six separate support vector machines were applied to mean wavelet power features extracted from multiunit activity to select one out of these six motions. The encoding scheme involved activating the movement associated with the highest decoder output using an electrode stimulation pattern previously calibrated to evoke that movement. Surface electrical stimulation was delivered as monophasic rectangular pulses at 50 Hz pulse rate and 500 µs pulse width, with stimulation intensity set to a piecewise linear function of decoder output. These results were extended to multi-joint reaching and grasping movement by Ajiboye *et al.* [42]: a linear decoder similar to a Kalman filter was used to map neuronal firing rates and high frequency power at electrodes in the hand area of the motor cortex to percent activation of stimulation patterns associated with elbow, wrist or hand movements. The researchers showed that a tetraplegic subject could perform multi-joint arm movements for point-to-point target acquisitions with 80–100% accuracy and volitionally reach and drink a mug of coffee.

One shortcoming of the above approaches is that continued electrical stimulation of muscles results in muscle fatigue, rendering the technique impractical for day-long use. An alternate approach to reanimation is to use brain signals to stimulate the spinal cord. Spinal stimulation may simplify encoding and control because it activates functional synergies, reflex circuits, and endogenous pattern generators. Capogrosso, Courtine, and colleagues [43] demonstrated the efficacy of brain-controlled spinal stimulation for hind limb reanimation for locomotion in paralyzed monkeys. They used a decoder based on linear discriminant analysis to predict foot-strike and foot-off events during locomotion. The encoder used this prediction to activate extensor and flexor "hotspots" in the lumbar spinal cord via epidural electrical stimulation to correctly produce the extension and flexion of the impaired leg.

*Restoring Motor and Cognitive Function*
One of the early pioneers exploring bidirectional BCIs for restoration of brain function was Jose Delgado [6] who designed an implantable BBCI called the stimoceiver that could communicate with a computer via radio. Delgado was the first to combine decoding with encoding to shape behavior: his decoding algorithm detected spindles in the amygdala of a monkey and for each detection, triggered stimulation in the reticular formation, which is associated with negative reinforcement. After six days, spindle activity was reduced to 1 percent of normal levels, making the monkey quiet and withdrawn. Unfortunately, efforts to extend this approach to humans to treat depression and other disorders yielded inconsistent results.

Delgado's work did eventually inspire commercial brain implants such as Neuropace's RNS system that detects onset of seizures using time- and frequency-based methods from brain surface recordings (ECoG) and stimulates the region where the seizure originates. Also inspired by Delgado's work is the technique of deep brain stimulation (DBS), a widely prescribed form of neurostimulation for reducing tremors and restoring motor function in Parkinson's patients. Current DBS systems are open-loop but Herron *et al.* have recently demonstrated closed-loop DBS [44] by triggering DBS based on movement intention, which was decoded as reduction in ECoG power in the low frequency ("mu") band over motor cortex.

*Enhancing Memory and Augmenting Brain Function*
Besides restoration of lost function, BBCIs can also be used for augmentation of brain function. Berger, Deadwyler and colleagues [45,46] have demonstrated that BBCIs implanted in the hippocampus of monkeys and rats can be used to enhance memory in delayed match-to-sample (DMS) and nonmatch-to-sample tasks. They first fit a multi-input/multi-output (MIMO) nonlinear filtering model to simultaneously recorded spiking data from hippocampal CA3 and CA1 during successful trials, with CA3 as input to the model and CA1 as output. The trained MIMO model was later used to decode CA3 activity and predict CA1 activity encoded as patterns of biphasic electrical pulses. Deadwyler *et al.* [46] showed that in the four monkeys tested, performance in the DMS task was enhanced in the difficult trials, which had more distractor objects or required information to be held in memory for longer durations. However, it is unclear how the approach could be used when the brain is not healthy such as in Alzheimer's patients [47] where simultaneous recordings from areas such as CA3 and CA1 for training the model in successful trials will not be available.

Nicolelis suggested several brain augmentation schemes based on BBCIs in his book [48], including direct brain-to-brain communication. He and his colleagues subsequently showed how rats can use brain-to-brain interfaces (BBIs) to solve sensorimotor tasks [49]: an "encoder" rat identified a stimulus and pressed one of two levers while its M1 cortex activity was transmitted to the M1 cortex of a "decoder" rat. The stimulation pattern was based on a Z score computed from the difference in the number of spikes between the current trial and a template trial. If the decoder rat made the same choice as the encoder rat, both rats were rewarded for the successful transfer of information between their two brains. Rao, Stocco and colleagues utilized noninvasive technologies to demonstrate the first human brain-to-brain interface [50,51,52]. The intention of a "Sender" who could perceive but not act was decoded from motor or visual cortex using EEG; this information was delivered via transcranial magnetic stimulation (TMS) to the motor or visual cortex of a "Receiver" who could act but not perceive. The researchers showed that tasks such as a video game [50] or "20 questions" [52] could be completed successfully through direct brain-to-brain collaboration (see [53,54] for other examples). More recently, brain-to-brain interfaces have been used to create a network of brains or "BrainNet" allowing groups of humans [55] or rats [56] to solve tasks collaboratively.

*Inducing Plasticity and Rewiring the Brain*
Hebb's principle for plasticity states that connections from a group A of neurons to a group B are strengthened if A consistently fires before B, thereby strengthening the causal relationship from A to B. Jackson, Mavoori and Fetz [57] demonstrated that such plasticity can be artificially induced in the motor cortex of freely behaving primates by triggering stimulation at a site B a few milliseconds after a spike was detected at site A. After two days of continuous spike-triggered stimulation, the output generated by site A shifted to resemble the output from B, consistent with a strengthening of any weak synaptic connections that may have existed from neurons in A to neurons in B. Such an approach could be potentially quite useful for neurorehabilitation by rewiring the brain for restoration of motor function after traumatic brain injury, stroke or neuropsychiatric disorders such as depression and PTSD. Along these lines, Guggenmos, Nudo, and colleagues [58] have shown that the approach can be used to improve reaching and grasping functions in a rat after traumatic brain injury to the rat's primary motor cortex (caudal forelimb area). Their approach involves creating an artificial connection between the rat's premotor cortex (rostral forelimb area or RFA) and somatosensory cortex S1 and for each spike detected by an electrode in RFA, delivering an electric pulse to S1 after 7.5 milliseconds. All of these prior approaches have relied on 1-to-1 spike-to-stimulation-pulse protocols, leaving open the question of how the approach can be generalized to induction of goal-directed multi-electrode plasticity.

**Towards A Unifying Framework: Neural Co-Processors based on Deep Learning**
A major limitation of current BBCIs is that they treat decoding and encoding as separate processes, and they do not co-adapt and jointly optimize a cost function with the nervous system. We propose that these limitations may addressed using a "neural co-processor" as shown in Figure 1. The approach uses two artificial neural networks, a neural co-processor (NCP) network and an emulator network, combined with a new type of deep learning that approximates *backpropagation through both biological and artificial networks*.

Suppose the goal is to restore movement in a stroke or spinal cord injury (SCI) patient, e.g., to enable the hand to reach a target location (see Figure 1). The NCP is a multi-layered recurrent neural network that maps neural activity patterns from a large number of electrodes in areas A1, A2, etc. (e.g., movement intention areas spared by the stroke or SCI) to appropriate stimulation patterns in areas B1, B2, etc. (e.g., intact movement execution areas in the cortex or spinal cord). When the subject forms the intention to move the hand to a target (e.g., during a rehabilitation session), the NCP maps the resulting neural activity pattern to a stimulation pattern. Unfortunately, to train the NCP, we do not have a set of "target stimulation patterns" that produce the intended movements. However, for any stimulation pattern, we can compute the error between the resulting hand movement and the target. How can this behavioral error be translated and backpropagated through the NCP network to generate better stimulation patterns?

We propose the use of an emulator network (EN) that emulates the biological transformation between stimulation patterns and behavioral output. The EN is a deep recurrent neural network whose weights can be learned using standard backpropagation from a dataset consisting of a large variety of stimulation (or neural activity) patterns in areas B1, B2, etc. and the resulting movements or behavior. After training, the EN acts as a surrogate for the biological networks mediating the transformation between inputs in B1, B2, etc. and output behavior.

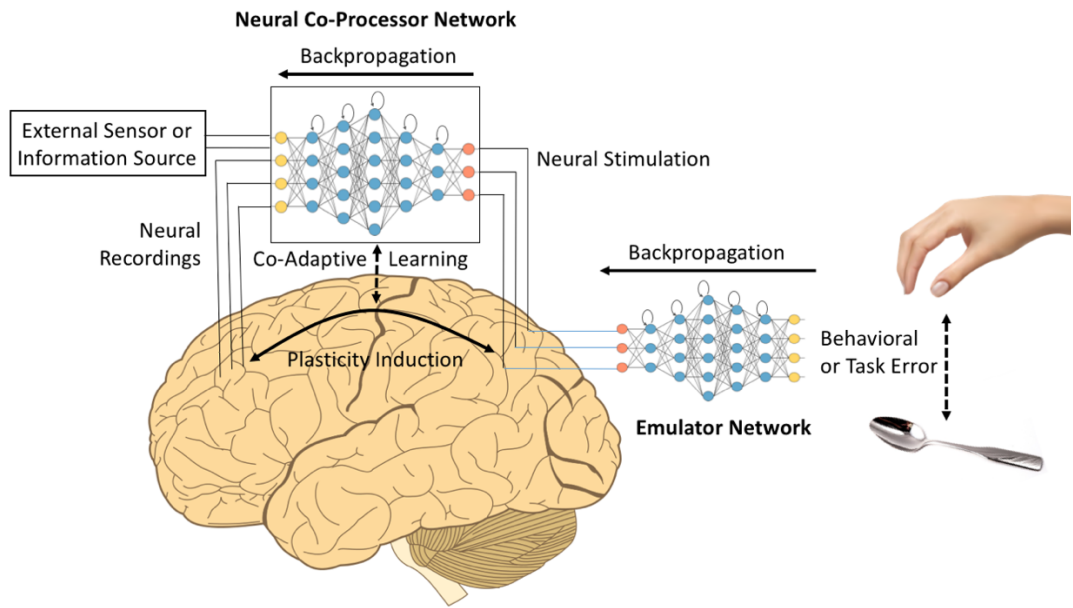

**Figure 1: Neural Co-Processor for the Brain for Restoring and Augmenting Function.** A deep recurrent artificial network is used to map input neural activity patterns in one set of regions to output stimulation patterns in other regions ("Neural Co-Processor Network" or NCP network). The NCP's weights are optimized to minimize brain-activity-based error (between stimulation patterns and target neural activity patterns when known), or more generally, to minimize behavioral error or task error using another network, an emulator network. The emulator network is also a deep recurrent network that is pre-trained through backpropagation to learn the biological transformation from stimulation or neural activity patterns at the stimulation site to the resulting output behaviors. During NCP training, errors are backpropagated through the emulator network to the NCP to adapt the NCP's weights but not the emulator network's weights. The trained NCP thus produces optimal stimulation patterns that minimize behavioral error, thereby creating a goal-directed artificial information processing pathway between the input and output regions. The NCP also promotes neuroplasticity between weakly connected regions, leading to neural augmentation or targeted-rehabilitation. External information from artificial sensors or other information sources can be integrated into the NCP's information processing as additional inputs to the neural network. The example here shows the NCP creating a new information processing pathway between visual cortex and the frontal lobe: this could be useful for restoring function after injury or stroke in an intermediate cortical area, or for boosting visual perception and visual decision making. A similar NCP could be used to transform movement intentions in the prefrontal cortex to appropriate movement-related stimulation patterns delivered to spinal neurons below the injury for restoration of movement and rehabilitation after spinal cord injury.

With the help of a trained EN, we can train the weights of the NCP network to produce the optimal stimulation patterns for minimizing behavioral error (e.g., error between current hand position and target location). For each neural input pattern X (e.g., movement intention) that the subject produces in A1, A2 etc., the NCP produces an output stimulation pattern Y in B1, B2 etc., which results in a behavior or movement Z.

The error E between actual movement Z and the intended movement target Z' is first backpropagated through the EN *but without modifying its weights*. We continue to backpropagate the error through the NCP network, this time modifying the NCP's weights. In other words, the behavioral error is backpropagated through a concatenated NCP-EN network but only the NCP's weights are changed. This allows the NCP to progressively generate better stimulation patterns that enable the brain to better achieve the target behavior, thereby resulting in a co-adaptive BBCI.

Furthermore, by repeatedly pairing patterns of neural inputs with patterns of output stimulation, the NCP promotes neuroplasticity between connected brain regions via Hebbian plasticity. Note that unlike previous plasticity induction methods [57,58], the plasticity induced spans multiple electrodes and is goal-directed since the NCP is trained to minimize behavioral errors. After a sufficient amount of coupling between regions X and Y, neurons in region X can be expected to automatically recruit neurons in region Y to achieve a desired response (such as a particular hand movement). As a result, in some cases, the NCP may eventually be no longer required after a period of use and may be removed once function is restored or augmented to a satisfactory level.

Finally, we note that NCPs can be extremely useful tools for testing new computational models of brain and nervous system function [31]. Rather than using traditional artificial neural networks in the NCP in Figure 1, one could use more realistic cortical models such as networks of integrate-and-fire or Hodgkin-Huxley neurons, along with biological learning rules such as spike-timing dependent plasticity rather than backpropagation. The challenge and a critical test for putative computational models of the nervous system would then be: can the model successfully interact with its neurobiological counterpart and be eventually integrated within the nervous system's information processing loops?

**Conclusions**
The ability to simultaneously decode neural activity from one region and stimulate another confers on BBCIs tremendous versatility, opening the door to applications ranging from prosthetic control and reanimation of paralyzed limbs to plasticity induction and enhancement of brain function. Promising results have been obtained in animal models and in some cases, humans but mostly under laboratory conditions.

To transition to real-world conditions, BBCIs must co-adapt with the nervous system and jointly optimize behavioral cost functions. We introduced the concept of a neural co-processor which uses artificial neural networks that jointly optimize behavioral error functions with biological neural networks. A trained emulator network is used as a surrogate for the biological network producing the behavioral output. Behavioral errors are backpropagated through the emulator network to the co-processor network which adapts its weights to minimize errors and delivers optimal stimulation patterns for specific neural input patterns. We illustrated how a neural co-processor could be used to improve motor function in a stroke or spinal cord injury patient. Such co-processors have not yet been validated in animal models or humans, but if successful, they could potentially be applied to modalities other than movement such as:

- Mapping inputs from one memory-related area to another to facilitate or restore access to particular memories (e.g., in memory loss) or to unlearn traumatic memories (e.g., in PTSD),
- Mapping inputs from one sensory area to another to restore or augment sensation and perception,
- Connecting areas involved in emotion processing to augment or rehabilitate emotional function, and
- Augmenting the brain's knowledge, skills, information processing, and learning capabilities with deep artificial neural networks.

**Acknowledgments**

This work was supported by NSF grants EEC-1028725 and 1630178, CRCNS/NIMH grant no. 1R01MH112166-01, and a grant from the W. M. Keck Foundation. The author would like to thank Eb Fetz, Chet Moritz, Andrea Stocco and James Wu for discussions related to topics covered in this article.